# Self-folding Self-replication

**Ralph P. Lano**
Technische Hochschule Nürnberg - Georg Simon Ohm
Keßlerplatz 12, 90489 Nürnberg, Germany
ralph.lano@th-nuernberg.de

**Abstract**
Inspired by protein folding, we explored the construction of three-dimensional structures and machines from one-dimensional chains of simple building blocks. This approach not only allows us to recreate the self-replication mechanism introduced earlier, but also significantly simplifies the process. We introduced a new set of folding blocks that facilitate the formation of secondary structures such as α-helices and β-sheets, as well as more advanced tertiary and quaternary structures, including self-replicating machines. The introduction of rotational degrees of freedom leads to a reduced variety of blocks and, most importantly, reduces the overall size of the machines by a factor of five. In addition, we present a universal copier-constructor, a highly efficient self-replicating mechanism composed of approximately 40 blocks, including the restictions posed on it. The paper also addresses evolutionary considerations, outlining several steps on the evolutionary ladder towards more sophisticated self-replicating systems. Finally, this study offers a clear rationale for nature's preference for one-dimensional chains in constructing three-dimensional structures.

**Keywords:** self-replication, self-assembly, protein folding, universal constructor, artificial life, nanorobots, programmable matter, computational origami

.

## Introduction

Interestingly living cells produce three-dimensional structures by folding of one-dimensional chains of amino acids. These chains are called peptides if they are short, or polypeptides if they are long. A protein consists of one or more of these peptides. One also calls peptides primary (1D), polypeptides secondary (2D), and the proteins tertiary (3D) structures. If a protein is composed of more than one polypeptide, it is referred to as a quarternary structure [1,2]. In a previous study we demonstrated a simple mechanical self-replication system [3]. In this study, we show that a similar mechanism can be implemented using a technique inspired by protein folding.

### Block Types

We begin with our basic building blocks, which are almost the same as those introduced previously [3]. Figure 1 recalls the five different block types: temporary blocks (Fig. 1(a)), building blocks (Fig. 1(b)), blocks for movement (Fig. 1(c)), blocks for gluing (Fig. 1(d)), and blocks required for sorting (Fig. 1(e)).

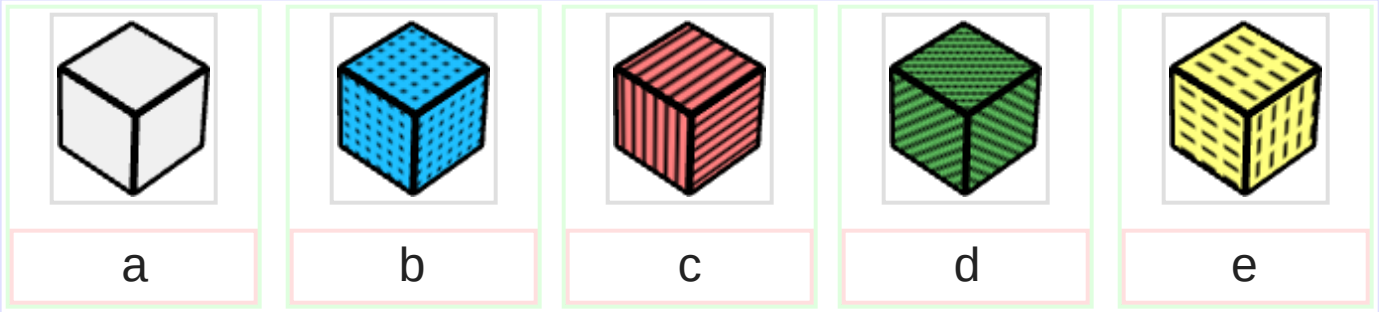

Figure 1: Standard block types

In addition to these standard block types, we introduced new folding block types, as shown in Fig. 2. These blocks exhibit special folding properties. All blocks have unique markings, which allow us to uniquely identify them through a mechanism similar to that indicated in [3]. Essentially, there is an unlimited supply of them, and no assumptions are made regarding the nature or size of these blocks, except that they are relatively small.

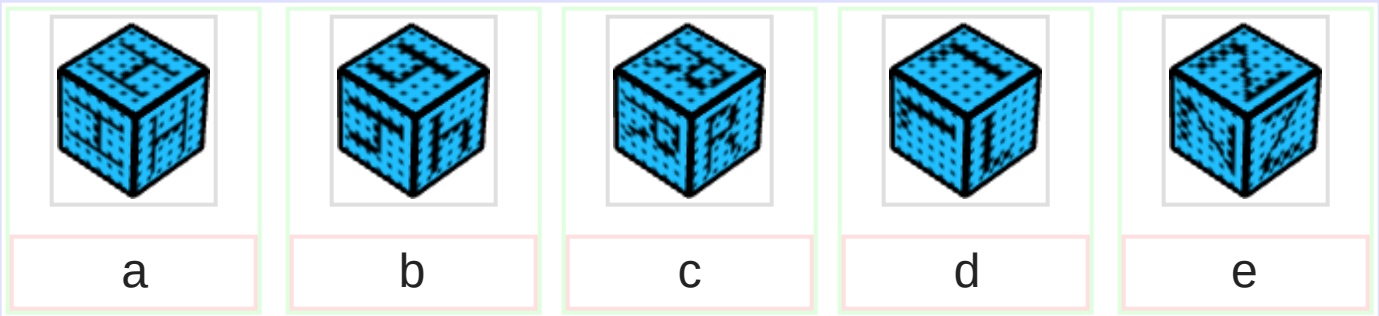

Figure 2: Folding block types

For descriptive purposes, we used the same nomenclature introduced in [3]. To describe a particular block type we use 'b' for Simple block, 'd' for Dissolvable block, 's' for Sorter block, 'G' for Gluer block, and 'M' for Mover block. New are the folding block types, and we use the abbreviations 'H' for hinge (Fig. 2(a)), 'h' for anti-hinge (Fig. 2(b)), 'R' for right rotation (Fig. 2(c)), 'L' for left rotation (Fig. 2(d)), and 'Z' for rotation by 180 degrees around the axis (Fig. 2(e)). Only the Gluer and Mover blocks require additional information for their descriptions, which are direction and timing. In contrast to [3], where we had to describe positions in a three-dimensional space, here we only describe one-dimensional structures. For instance, "b__H__b__b__b__" describes a chain of blocks, which is read from left to right. It begins with a Simple block, followed by a Hinge block and then three more Simple blocks.

## Chains

Figure 3a shows that the markings at the bottom of each block type can not only be used to identify the block type, but also to carry directional information by choosing a proper encoding scheme, as discussed in [3]. If one chooses the glue mechanism from [3], where blocks can be glued on all sides, that is all that is needed. Figure 3b shows, how the blocks can be glued together to form chains.

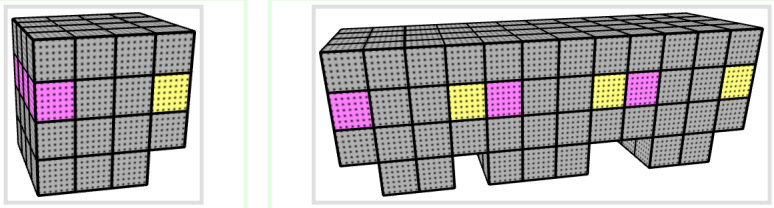

a b

Figure 3: Markings and directional information

Previously, the ablility to glue blocks from all sides was a key requirement. However, this severe restriction is no longer required when working with chains. This requirement can be lifted, as indicated in Fig. 3(a): blocks only need to be glued together at two locations, which are marked in magenta and yellow. Note that these locations are off-center, and hence uniquely define the 3D direction. This is a significant simplification over the old gluing mechanism. However, as we will see later, this has consequences for the Copier mechanism. Let us remark that Fig. 3 is one possible way blocks can look like when zoomed in, but it is definitely not the only way.

.

## Folding

To understand the function of the folding blocks, we consider a few examples. Figure 4 shows how the folding using the Hinge block ('H') works. Fig. 4(a) shows the chain before folding, which is also called the primary structure in protein folding. When constructing the chain, the description "b__H__b__b__b__" is read from left to right. It should be noted that there are two fixed directions here: one is the direction of addition of new blocks (the x-direction) and the other is the upward direction (the z-direction). Some time after bonding, the Hinge block angles upwards by 90 degrees, as shown in Fig. 4(b). The Hinge block can only angle upwards relatively. It does this only once and then remains permanently in this position. The Anti-Hinge block ('h') angles downwards by 90 degrees.

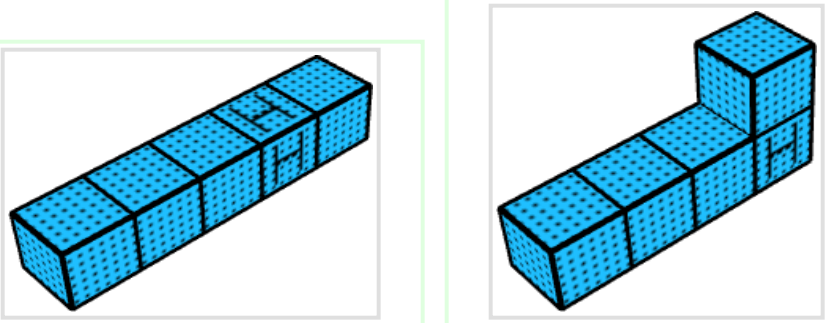

a b

Figure 4: Folding of the Hinge block

Figure 5 shows the function of the different rotating blocks. The Hinge block can only angle in a fixed direction. However, one would like to be able to angle in all possible directions, and for this, rotation blocks are required. The rotation blocks rotate the relative direction of the chain either to the right (Fig. 5(b)), by 180 degrees (Fig. 5(c)), or to the left (Fig. 5(d)). Rotation blocks can also be used together with Gluer and Mover blocks, which means that only one direction is needed for the Gluer blocks instead of six. For Mover blocks, two directions are required instead of six: one in the direction of the chain and one perpendicular to it. Overall, this effectively reduces the number of Mover block types required to a third.

This is important to understand: although introducing five new block types seems to worsen our encoding problem, but because we no longer need so many Gluer and Mover block types, in the end this is actually beneficial. In the old scenario, we potentially had 74 different block types, whereas in the new scenario, this was reduced to 31. In addition, we want to point out that the functionality of the 'h' can be achieved by combining 'H' together with 'Z'. And that assuming we only had 'L' at our disposal, we could get the effect of a 'Z' be combining two 'L', and the effect of an 'R' could be achieved via three 'L'.

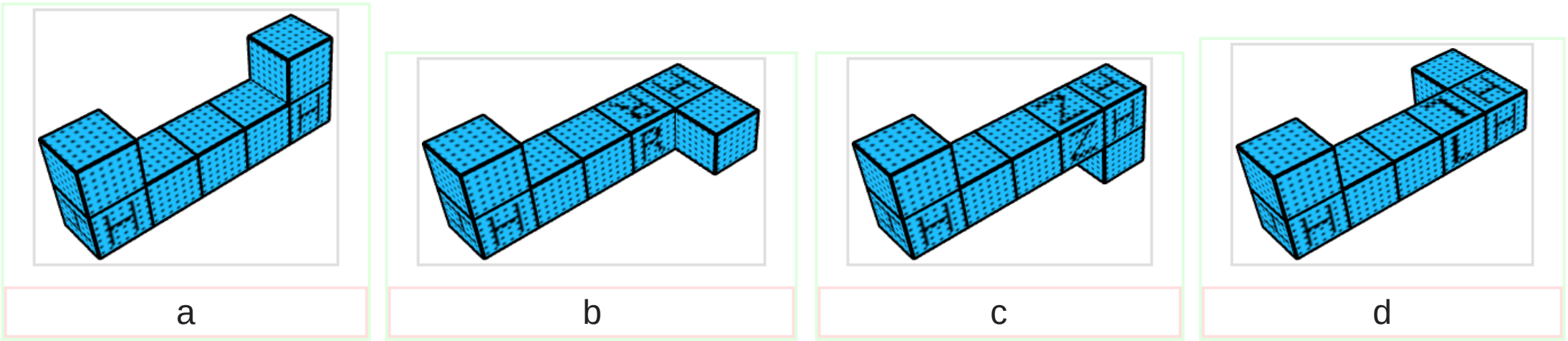

Figure 5: Rotating block types

Figure 6 shows the algorithm used for folding. This algorithm processes a string of blocks, represented by three-character chunks, to generate a 3D positional array of the folded protein structure. It initializes an empty hashmap to store coordinates and blocks, iterating through the string describing the chain, in increments of three characters. Depending on the first character of each chunk, the hash map may be shifted to the right or rotated along the X or Z axes by specific degrees. Chunks starting with "b", "d", "M", "G", "1", or "2" require no rotation. After handling the rotations, the algorithm stores the chunk at coordinate (0, 0, 0) in the hash map. Finally, the hash map is converted into a 3D positional array that represents the folded chain.

```
Algorithm Folding
Input: code - a string in the form "b__R__H__M__Z__b__H__b__H__"
Output: s3d - a 3D array of strings

Initialize an empty hashmap hm to store coordinates as keys and
corresponding 3-character strings as values
For each x from 0 to length(code) - 1 by 3:
  s ← substring(code, x, x + 3)
  hm ← shiftRight(hm)
  If s starts with "b" or s starts with "d" or s starts with "M" or
     s starts with "G" or s starts with "1" or s starts with "2":
    Do nothing
  Else if s starts with "R":
    hm ← rotateAlongX(hm, -90)
  Else if s starts with "L":
    hm ← rotateAlongX(hm, +90)
  Else if s starts with "Z":
    hm ← rotateAlongX(hm, 180)
  Else if s starts with "H":
    hm ← rotateAlongZ(hm, 90)
  Else if s starts with "h":
    hm ← rotateAlongZ(hm, -90)
  Else:
    print("Unknown symbol")
  hm[ (0, 0, 0) ] ← s
  s3d ← convertHMTo3DStringArray(hm)
```

Figure 6: Algorithmic description of folding mechanism

.

## Secondary Structures: α-Helix and β-Sheets

Using this mechanism, we can build simple one-dimensional and two-dimensional structures. Figure 7 shows simple, repeating chains that form left-handed helices (Fig. 7(a)), right-handed helices (Fig. 7(b)), or zigzag structures (Fig. 7(c)). In protein folding, the former are also referred to as α-helices. For example, the use of left-turning blocks leads to left-handed helices. These secondary structures can effectively take on support and structural functions.

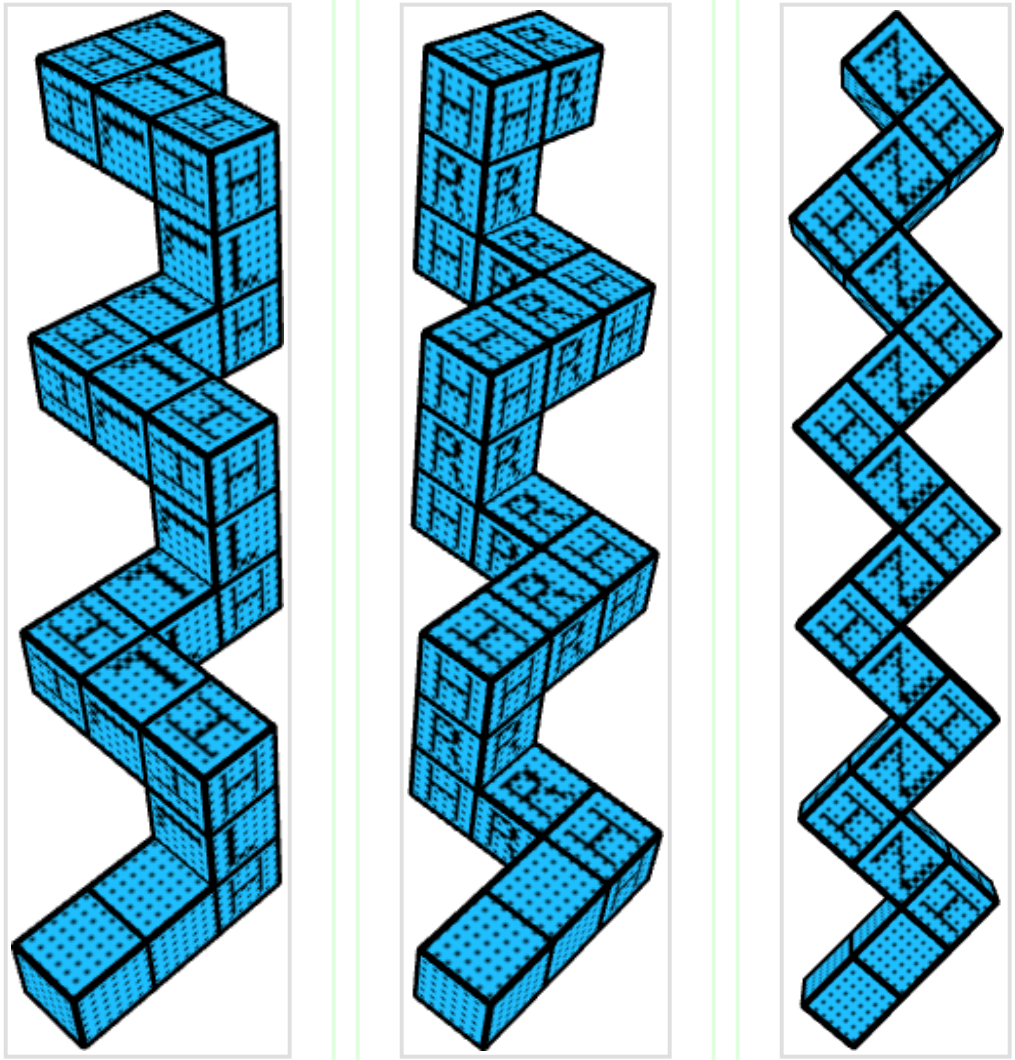

Figure 7: Simple α-helices

Figure 8 shows simple planar structures, which are also referred to as β-sheets in protein folding. They can be extended or widened arbitrarily by repeatedly inserting Simple blocks. These secondary structures can effectively take on support, separation, transport, and structural functions.

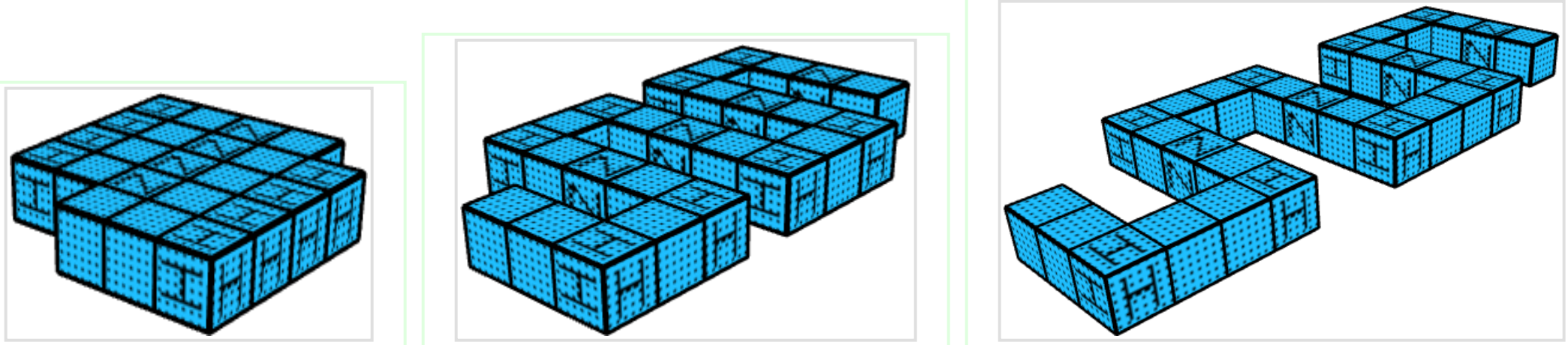

Figure 8: Simple β-sheets

.

## Sticks, Codons and tRNA*

For our purposes, important simple two-dimensional structures are 3-Sticks, codons and tRNA*s. Figure 9 shows 2-codons and 2-tRNA* produced using the folding technique. Looking at the upper extrusion of the codons, one can see the binary numbers 00 (Fig. 9(a)), 01 (Fig. 9(b)), 10 (Fig. 9(c)), and 11 (Fig. 9(d)), with the least significant bit on the right side. For tRNA*, the extrusion is at the bottom, and the binary numbers can also be seen here. As long as we allow for gluing at all sides this is just fine.

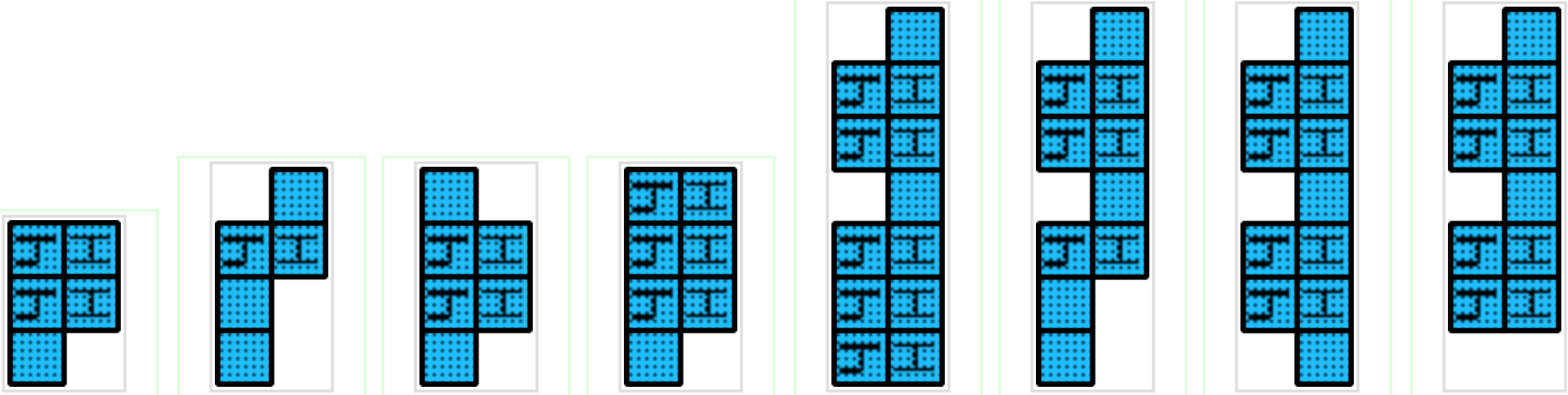

Figure 9: Codons and tRNA*

.

## Tertiary Structures: Machines

After these preliminary studies, we would like to continue with the main part of this work, that is, the implementation of the machines from our previous work [3], using the folding technique. These machines consist of functional blocks and support structure. The central idea is to keep the functional blocks at their respective locations and orientations, while modifying the support structure such that it can be implemented via folding.

### Builder

It is instructive to see how to convert our old machines into the folding world, using the Builder as an example. As indicated, we want to keep the functional blocks (Movers, Gluers and Sorters) at the same relative positions with respect to each other. Figure 10(a) depicts the appearance of the oirignal Builder. In a first step, we remove all the supporting Simple blocks, and in a second step, we try to connect the remaining functional blocks through a single line. There is the restriction, that the "working area" of the machine is not obstructed. Figure 10(b) shows one possible choice of connecting the functional blocks through a single line while maintaining their location and not obstructing the working area.

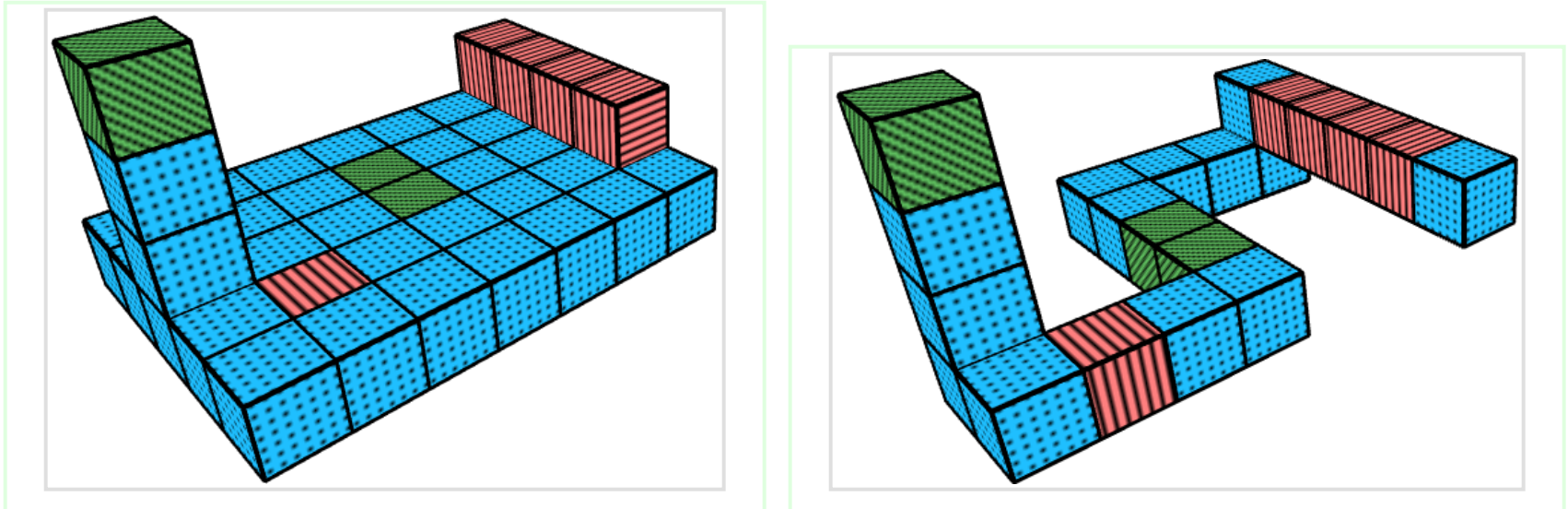

a b

Figure 10: Builder

In general this is the idea for converting the machines from [3] to the new folding technique: One begins with the original machine and focuses on the functional blocks, which should remain in the same relative position and point in the same direction. In an initial step, one tries to connect these active blocks through a line. In a second step, one then uses the hinge and rotational blocks to recreate this line. And in a third step, one pays attention to the proper orientation of the functional blocks, which is achieved with the help of inserting additional L-, R-, and Z-blocks. The result is shown in Fig. 11(a), it is functionaly completely equivalent to the original Builder.

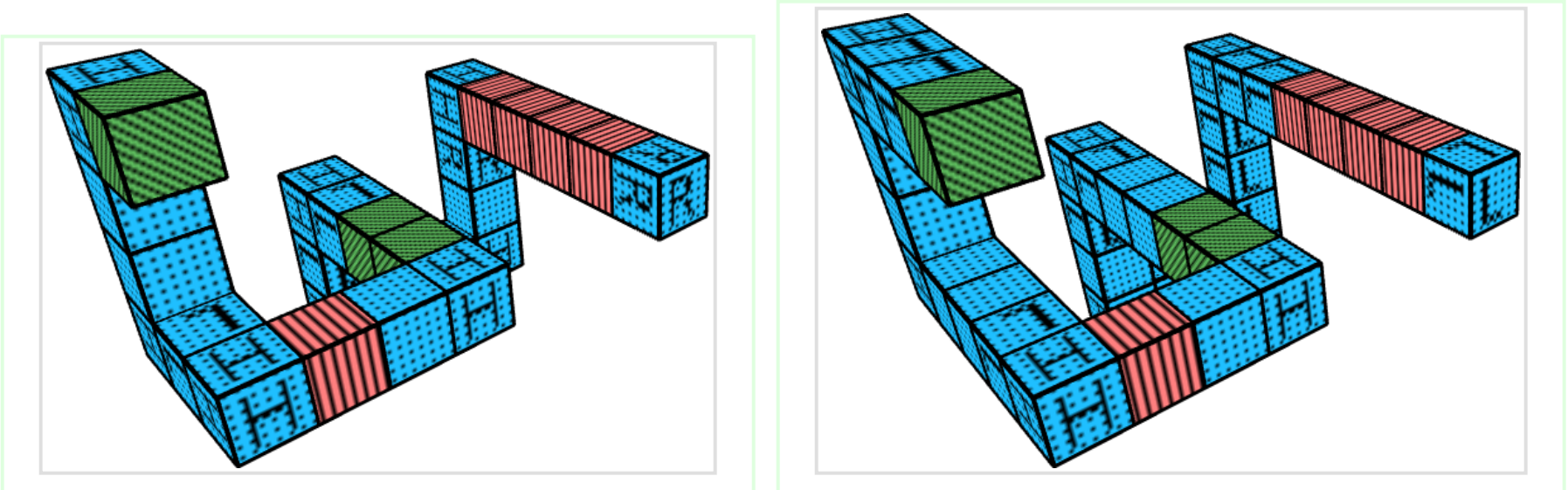

a b

Figure 11: Builder through folding

For coding efficiency reasons, the question as to whether the Builder can also be made only from H- and L-blocks, that is, without using R- and Z-blocks. Figure 11(b) shows that this is possible for the Builder. In this case we require six different block types, namely b, H, L, G, and two types of M blocks.

To describe the machines, we used a slightly modified version of the machine description language introduced in [3]. Instead of a three-dimensional description of the location of the blocks, a simple one-dimensional description is sufficient. Figure 12 shows the modified MDL uniquely describing the Builder shown in Fig. 11(a). This is read line-by-line starting from the upper left. The MDL description of all the structures and machines presented here can be found in the Appendix.

```
G2_H__b__b__H__L__H__M24b__H__G2_G2_L__H__L__H__b__b__H__b__
R__H__M50M50M50M50R__
```

Figure 12: MDL for Builder

The Builder created using the folding technique was 27 blocks long. This is to be compared with the 4*5*7, that is 140, blocks needed for the old Builder. Hence, the folding technique results in a reduction of building material by a factor of five, and the RNA needed to encode this information (i.e. the genome) is also recuded by a factor of five. This is to be expected, since in the old case, we encode a volume, O(n^3), whereas in the new case, we encode a line, O(n).

## Decoder

The original Decoder reads information from RNA by matching it with 2-tRNA and creates a stream of ordered blocks. The functionality of the Decoder is described in detail in [3]. It is straightforward to convert the original Decoder into its folding equivalent, as depicted in Fig. 13(a), following the procedure outlined above.

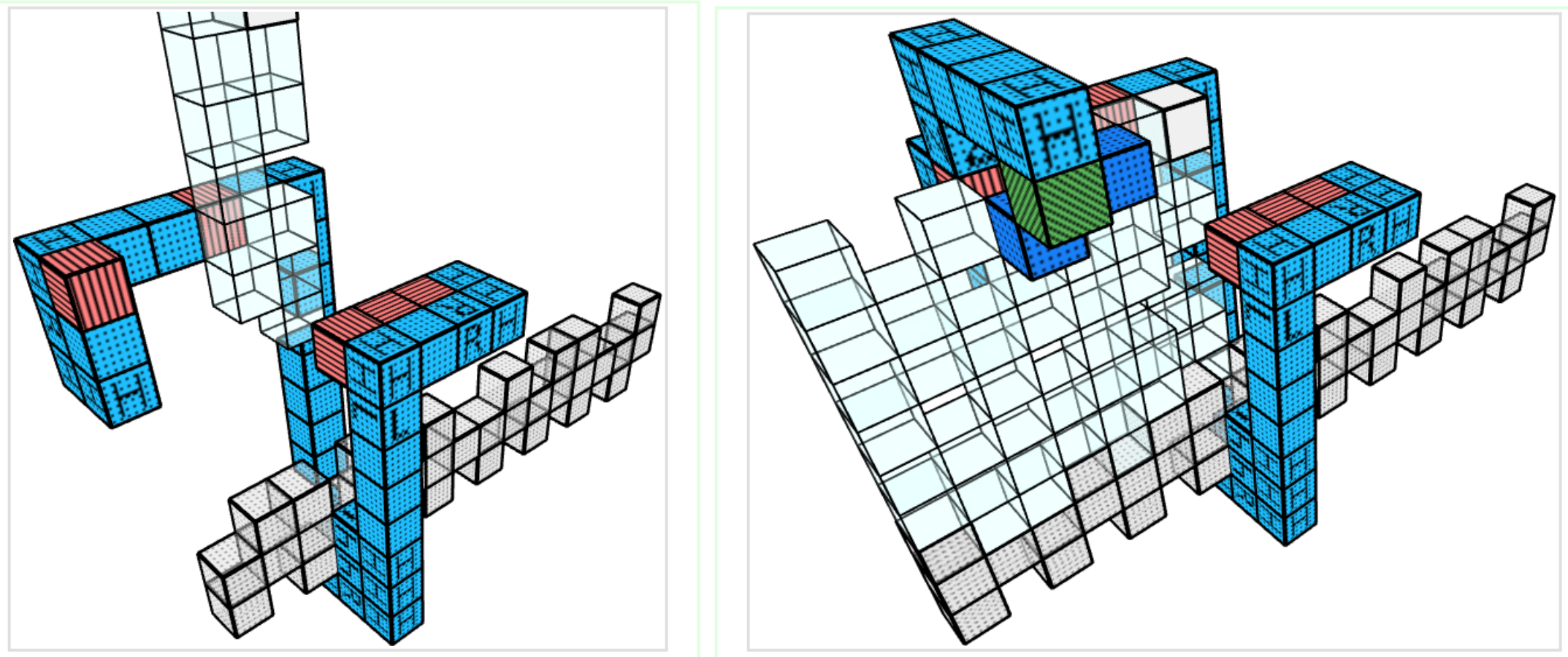

a b

Figure 13: Decoder without and with Gluer block

More interestingly, however, is a small modification of the original Decoder: by simply adding a Gluer block to the Decoder (Fig. 13(b)), the stream of ordered blocks leaving the Decoder is immediately glued together and turned into a chain, which then folds itself some time after gluing. Interestingly, this completely eliminates the need for the Builder. In the folding world, the Builder is no longer needed: the structures and machines assemble themselves through folding. This new Decoder somewhat resembles the functionality of the ribosome protein in the biological world.

Folding is a very powerful technique. To see this, let us revisit the overview diagram presented in [3], which no longer includes the Builder. Not only the sticks, codons, and tRNA* can be produced through folding, but as we will show in the following, all other machines as well.

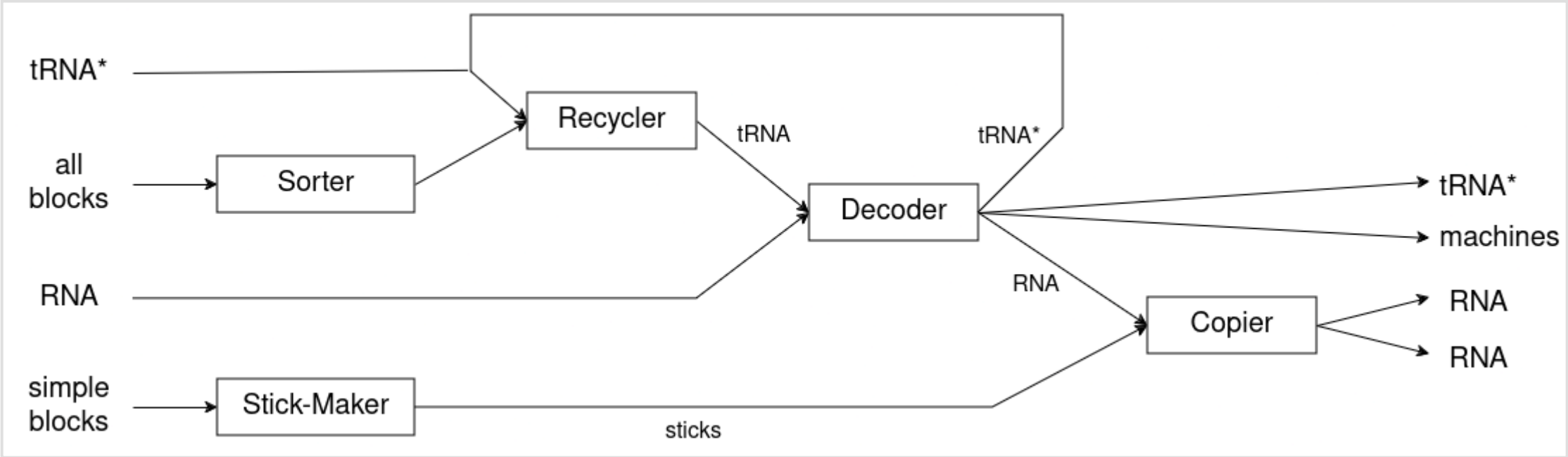

Figure 14: Overview of folding mechanism

.

## Recycler

The Recyclers take spent 2-tRNA* from the Decoder and refill them to create filled 2-tRNA. Figure 15 shows the front view of the Recyclers, where the shapes of the different tRNA* can be recognized. 2-tRNA* is matched and enters from the back. In addition, blocks sorted by the Sorter come from above into the Recyclers, and in this way, the empty tRNA* are filled. The details of this process are described in [3]. Modelling the Recyclers in the folding technique is somewhat of a challenge, but in the end is a straightforward process.

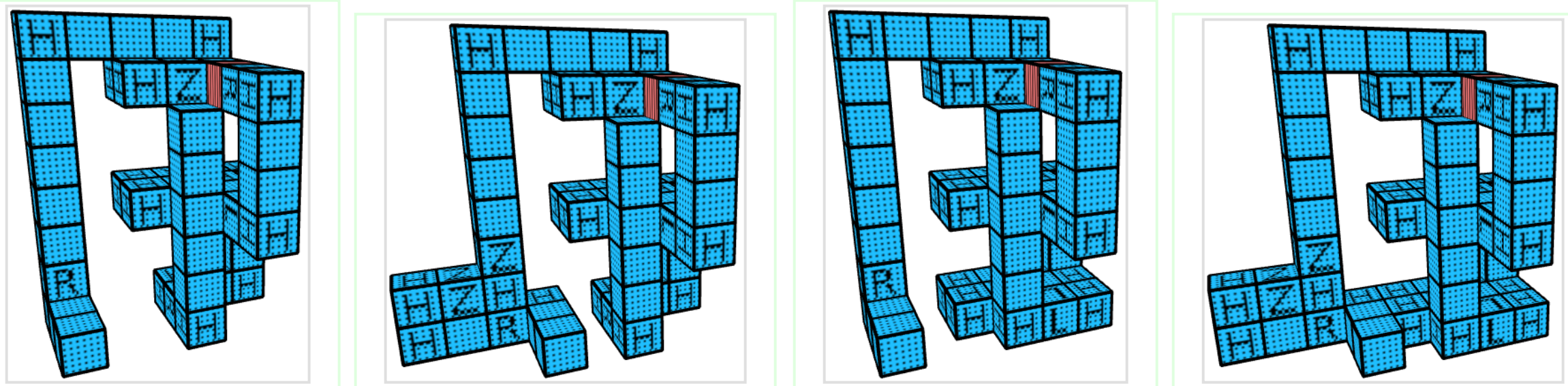

a b c d

Figure 15: Recyclers

.

## Sorter

The Recyclers need sorted blocks as input, which the Sorter machine provides. Sorter blocks are used to distinguish the different block types based on their respective unique markings. Figure 16 shows a single Sorter machine produced using the folding technique. There are three interesting observations here.

First, creating the whole Sorter will be a bit more complicated, but it is important to note, that in the construction shown, both ends, the beginning and end of the chain are available for continuation. This implies that several single Sorters can be connected.

Second, the Sorter shows an interesting use of Dissovable blocks (depicted in white) in the context of the folding technique. In Fig. 16, we see a Hinge block between Dissovable blocks. As long as the Dissovable blocks are not dissolved, the machine cannot start working. The first time the machine starts working, there is a Hinge block, where there should be none, resulting in a mistake in the first run. In addition, this assumes that the Dissovable blocks dissolve after the folding has finished.

Third, a big advantage that folding has, is that it requires only half of the number of Sorter blocks (depicted in yellow). As can be seen by comparing Fig. 16(a) and Fig. 16(b), the same Sorter block is used, however, because the loop has two different orientations, the Sorter block in Fig. 16(a) is mirrored with respect to Fig. 16(b).

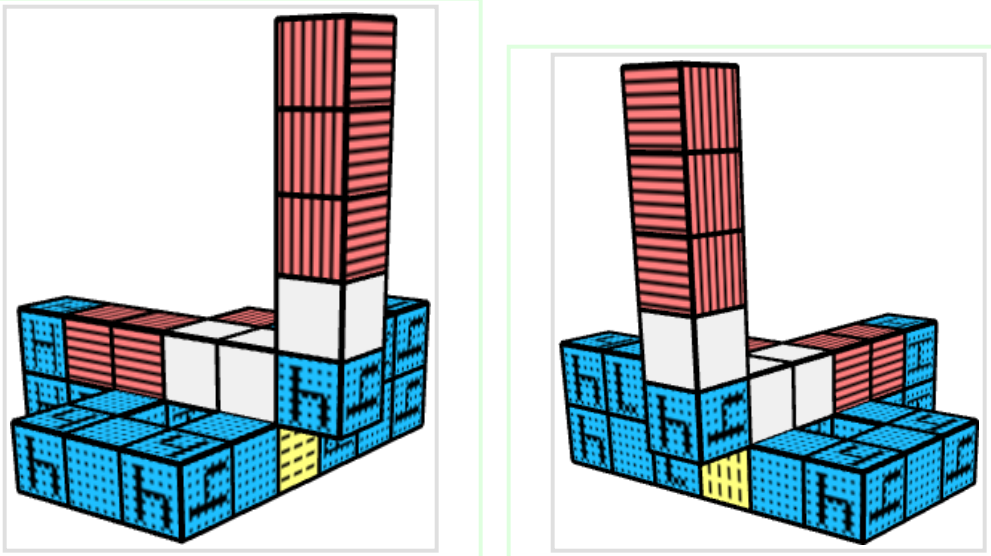

a b

Figure 16: Simple Sorter

.

## Stick-Maker and Copier

Figure 17 depicts a 3-Stick (Fig. 17(a)), the Stick-Maker (Fig. 17(b)), and the Copier (Fig. 17(c)) produced via the folding technique. The Stick-Maker creates 3-Sticks, which are then used by the Copier to produce a copy of the

RNA tape.

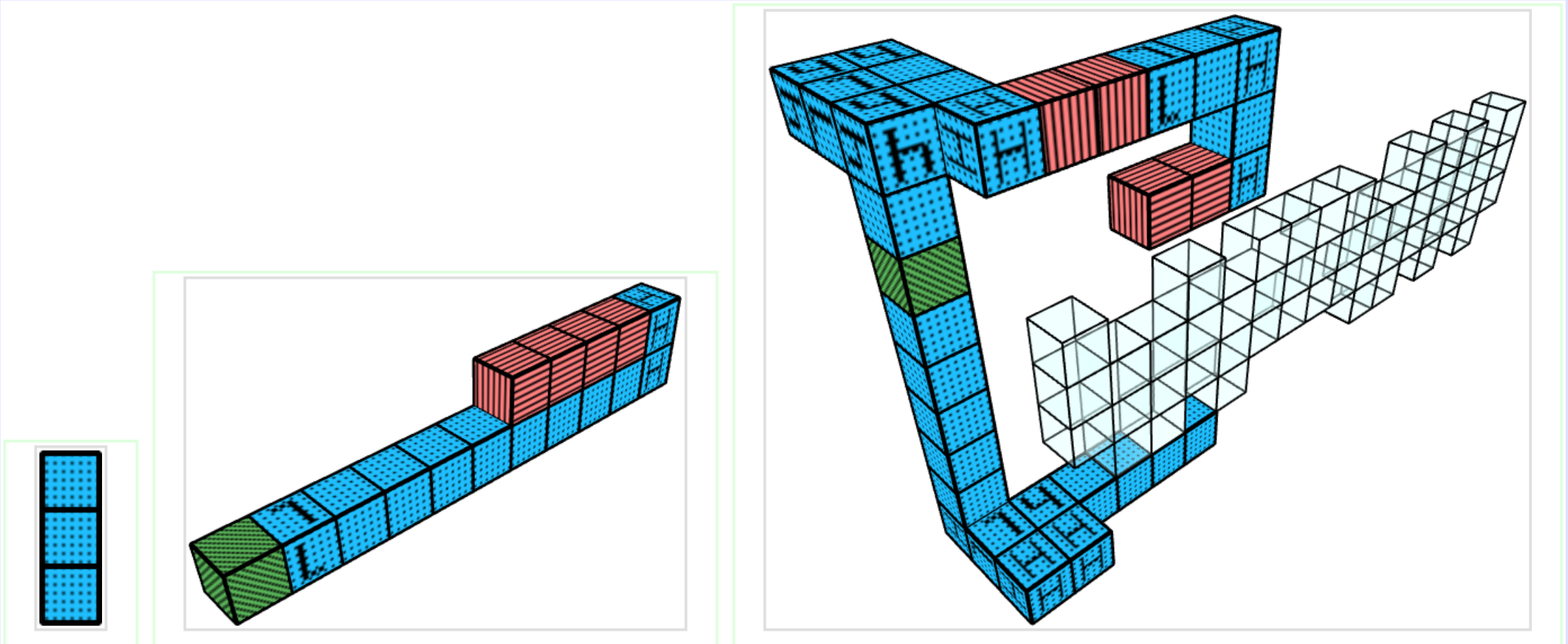

a b c

Figure 17: Stick-Maker and Copier

As we have already indicated above, there might be an issue with this copy process depending on the exact nature of gluing. If the glue mechanism is the same as in reference [3], that is gluing on all sides is possible, there will be no issues. However, if we had to use something like the restricted gluing mechanism introduced in Fig. 3, we have the following problem: in this mechanism the 3-Sticks ("b__b__b__") can only be glued at the top or bottom part. They could not be glued at their sides. However, this is a critical assumption for the Copier to work.

There are two solutions to this problem: the simple one would be to modify the glue properties of only the Simple blocks, such that they can also be glued on all sides. The second is to propose an alternative copying mechanism.

## Alternative Copy Mechanism

If we were to have a mechanism based on the restricted gluing, then neither the 3-Sticks (Fig. 17(a)) nor the normal 2-codons (Fig. 9(a)-Fig. 9(d)) can be glued together to form an RNA tape. However, if we introduce the codons shown in Fig. 18, RNA tapes can be formed. In this figure, the white blocks are not part of the codons but are indicators of the beginning and end of the chain.

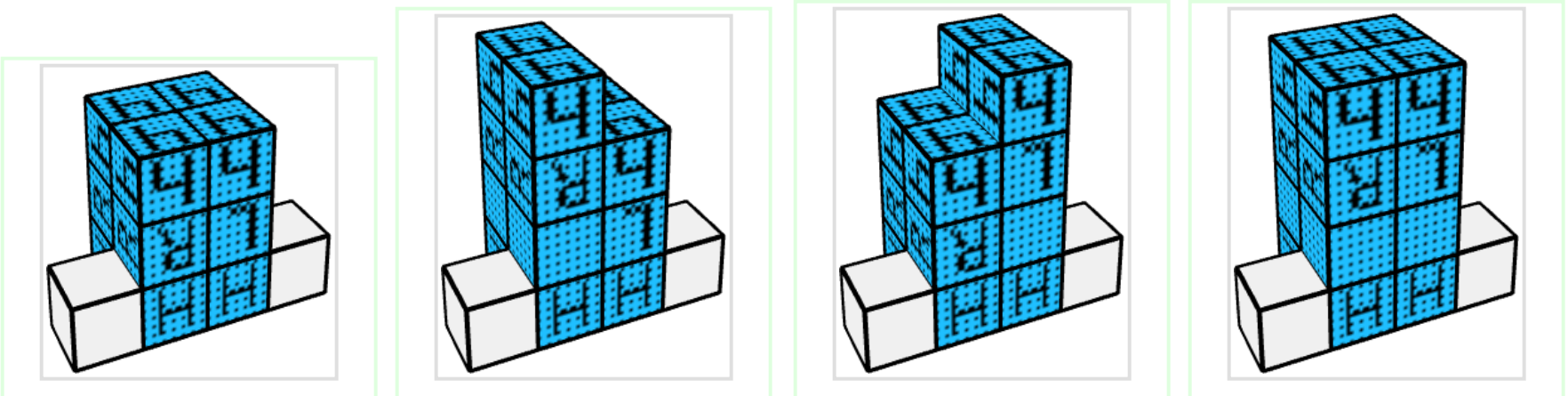

a b c d

Figure 18: Modified 2-codons

The copy mechanism is then slightly more complex. Figure 19 shows a Copier produced using the folding technique, which creates a negative copy of a given RNA tape. The RNA tape and sorted 2-codons are used as inputs. The feed forward for the RNA tape is achieved using a move-by-two conveyor (not shown). The RNA tape is fed from the left side and the 2-codons enter from above. Then, Mover blocks attempt to match the 2-codons with the RNA tape in the sequence from back to front. Through further feeding, the 2-codons are glued together using the Gluer block. As output, the Copier produces a negative copy of the original. Interestingly, the Copier can also make copies of an RNA tape composed of any n-codons, where n is even, without any modifications. If this process is repeated with the help of another similarly constructed Copier, a positive copy of the original can be produced.

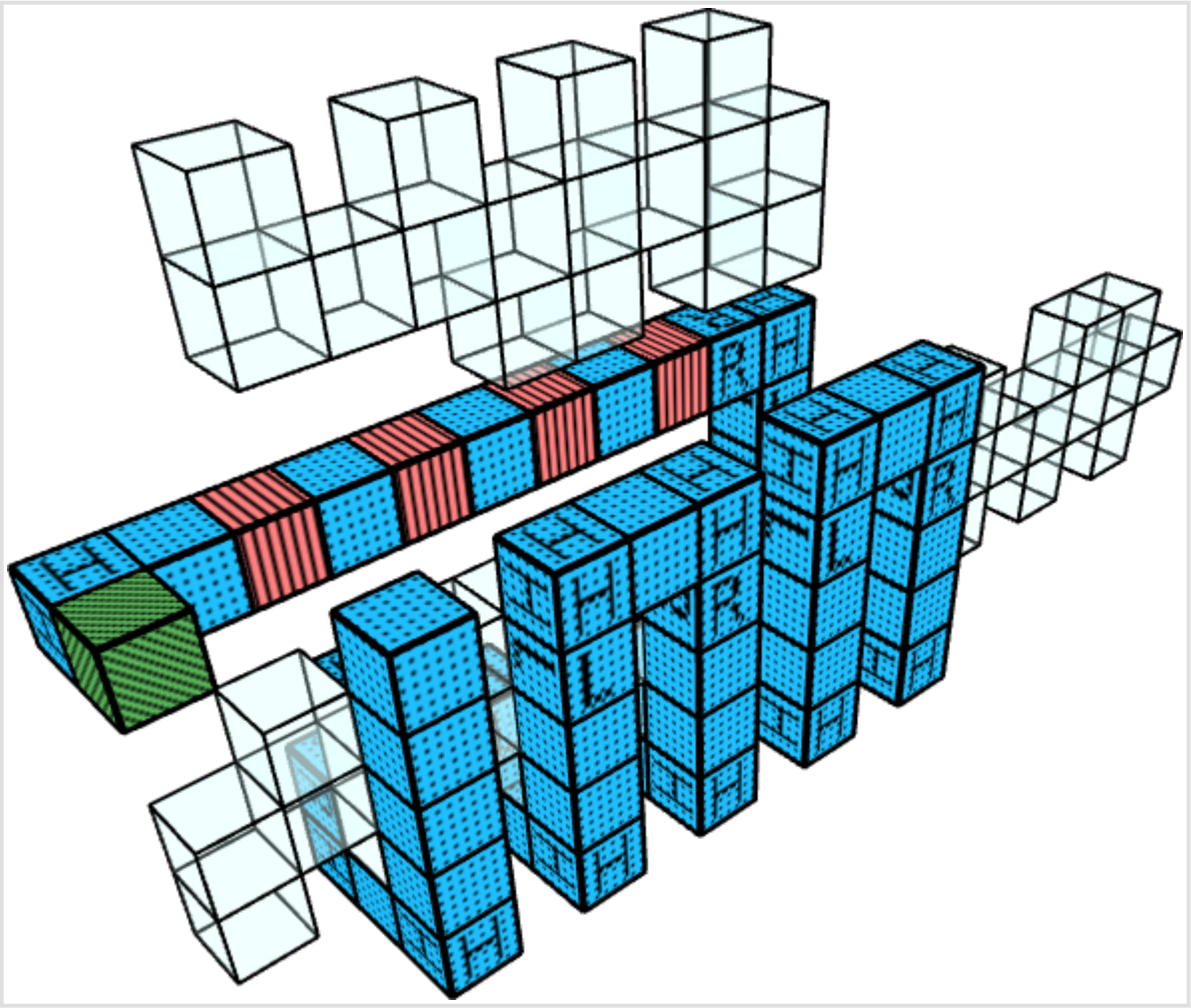

a

Figure 19: Negative Copier

Although significantly more complicated, this copy mechanism works in principle. However, one issue needs to be addressed: where do the modified 2-Codons come from? In our old scenario, the Stick-Maker was mass-producing 3-Sticks. We need a mechanism to mass-produce modified 2-Codons. One possible solution would be to have a dedicated Decoder, which repeatedly feeds the same RNA to only produce modified 2-Codons. However, this process is not very efficient.

.

## Quarternary Structures

From the structures and machines described above, it appears that there is no longer any need for the Dissovable blocks. The exception was the Sorter machine, where the use of the Dissovable blocks allowed us to form the Sorter out of a single chain of blocks. This is a useful technique. Assuming we had one large original RNA tape, holding the information for building all machines. How would we know where one machine ends and where the other begins? If the different machines were to be connected via Dissovable blocks, the problem of separation would be solved very easily. Furthermore, some of our machines are composed of several parts. Using Dissovable blocks these can be constructed from a single chain. In the following, we show other interesting applications of the use of Dissovable blocks, which may lead to the need for Dissovable block types with different dissolving timing properties.

.

### Track and Walker

Track and Walker are an example of two structures joined by a Dissolvable block. Figure 20 shows a Track (Fig, 8(b)) together with a Walker from reference [3]. Initially, they are connected via a Dissolvable block. Once this dissolves, the Walker can start moving. For instance, this can be used to bring material to a specific location. This is reminiscent of the Kinesin protein in the biological world.

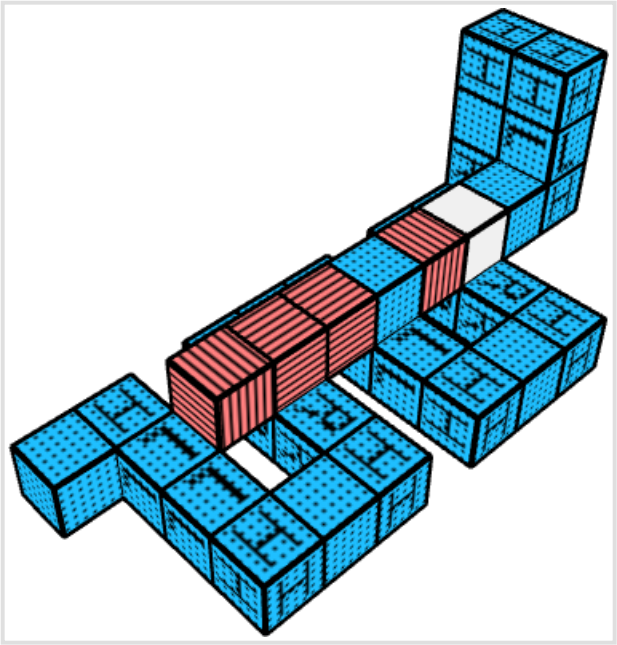

a

Figure 20: Track and Walker

## Retainer

The previous example shows how to allow movement in one direction. The Retainer (Fig. 21) is an example that allows movement in two directions. What we want is z-movement after the Walker reaches the end of the Track. The Retainer keeps the Walker connected to the Track. The two additional Movers want to move up in the z-direction, but as long as there is a Track, the Retainer will prevent it. Only at the end of the Track will the z-movers expand, and move the payload away from the Track. In this manner, material can be specifically moved to any point in the x-z plane.

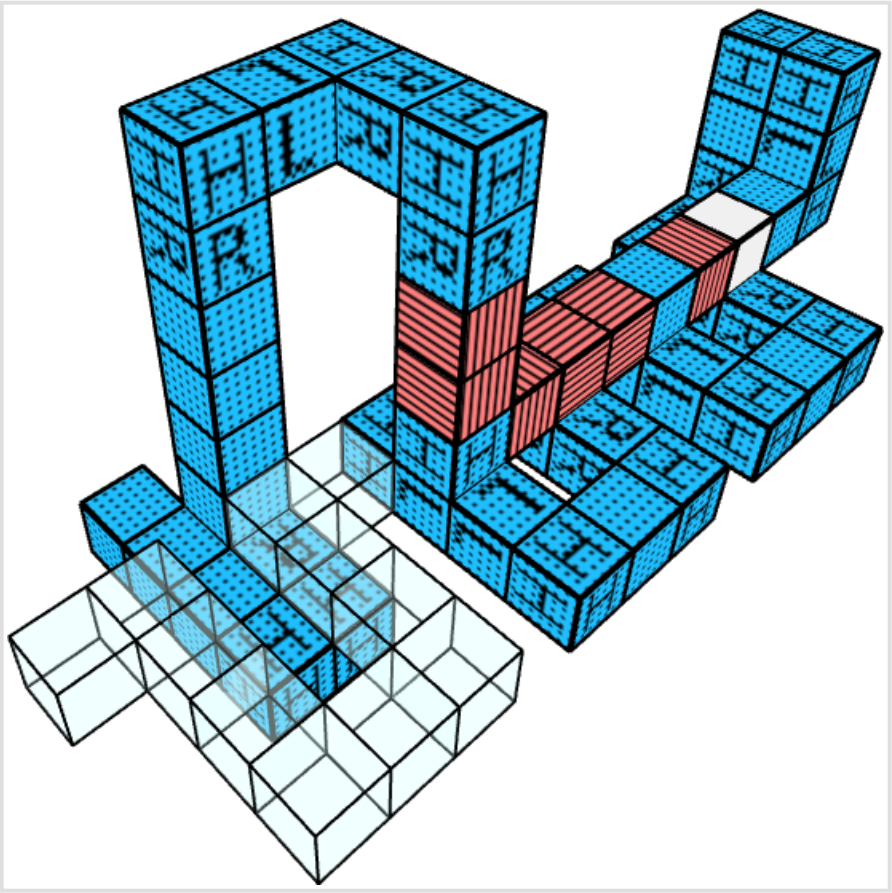

a

Figure 21: Track and Walker with Retainer

## Shuttle

Figure 22 shows how material can be continuously moved from one place to another, or back and forth. For this purpose, a Shuttle (FIg. 22(c)) is required, which effectively consists of two opposing Walkers. The Shuttle moves on a Track (FIg. 22(a)) such that only one of the two walkers is in the middle of the track (FIg. 22(b)). Consequently, the Shuttle moves in only one direction. At the ends of the track, two movers ensure that the Shuttle is pushed to the other side. This now places the other Walker in the middle, and the Shuttle moves in the opposite direction.

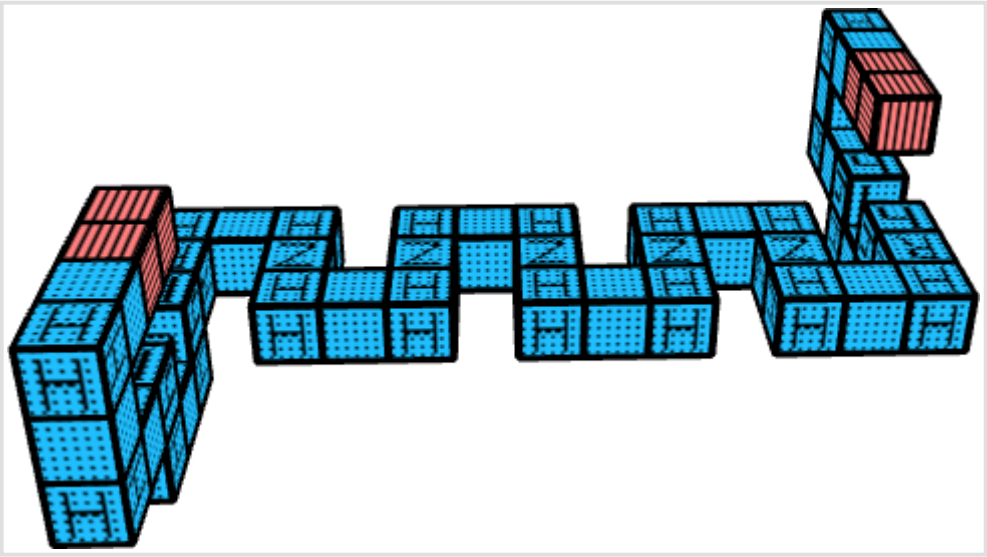

a

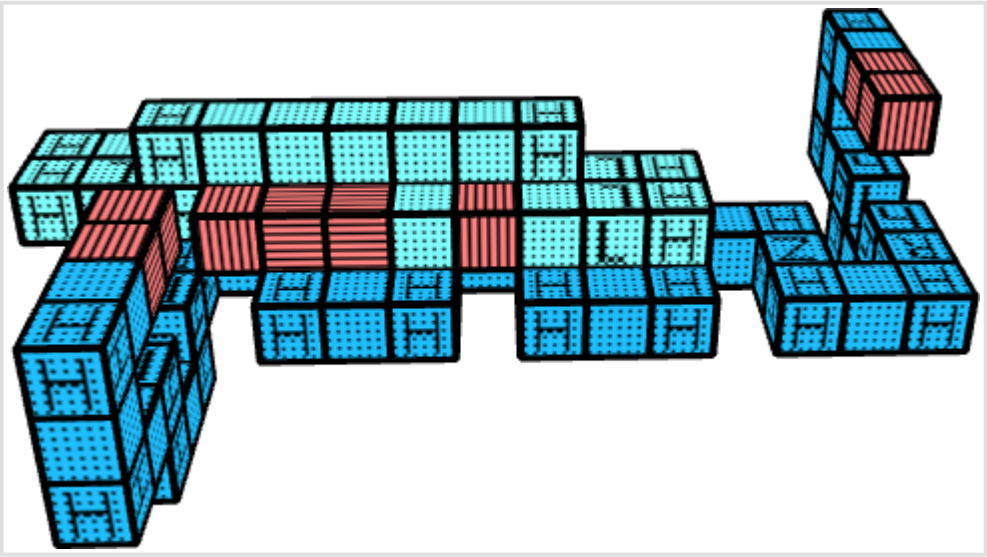

b

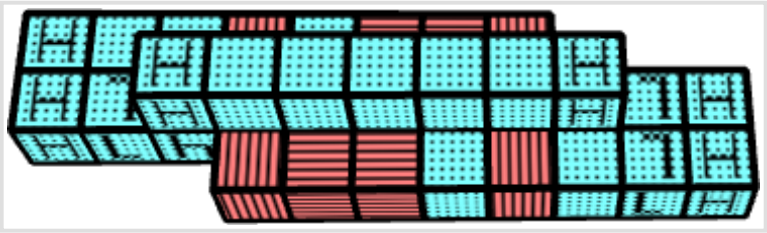

c

Figure 22: Track with Shuttle

### Timer

Another use of the Dissolvable block is as a timing device. Figure 23 shows the manner in which a certain time delay can be achieved using a Dissolvable block. Only after the Dissolvable block has dissolved can the two movers expand.

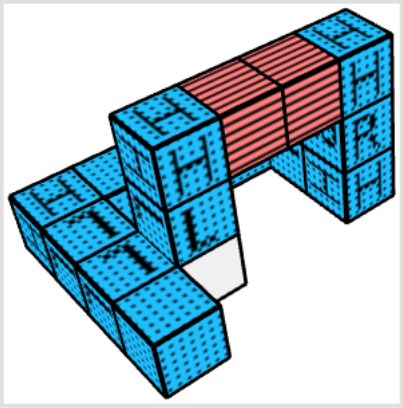

a

Figure 23: Timer

.

## Universal Copier-Constructor

When looking at the markings of the Sorter blocks and the encoding of the tRNA, one sees that the tRNA just translates between one and the other. The tRNA provides a one-to-one relationship between the encoding scheme and block type. Thus, looking at the anti-codon part of the tRNA and the markings of the Sorter blocks, one wonders that they serve a similar purpose: distinguishing block types. Hence, one wonders whether one needs tRNA at all?

Similarly, does one need the RNA tape at all? The machines kind of have the information built into them. We just need to unfold them and carefully read their markings. Hence, if one was to unfold them, copy them, and then fold them again, one would not need an RNA tape at all. Depending on the folding, this may be hard or impossible. However, if one were to make the copy before the folding occurred, there would be no need for unfolding. Hence, if one were to keep an unfolded copy of the machine, it would serve as an RNA tape. Thus, the only difference between RNA tape and machine is the folding. Copying and building are the same process. All that building is, is an event that triggers the folding. Figure 24 then shows this simplest self-replicating physical system imaginable, and in the following, we discuss how an implementation might be possible, and what some of the constraints on the system and its constituents are.

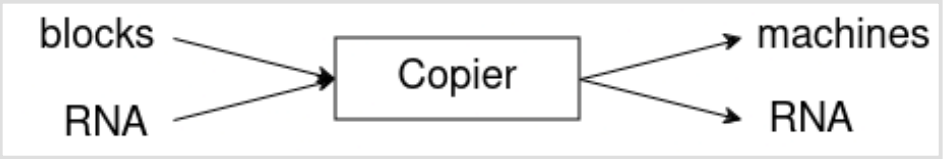

Figure 24: Overview

.

## Assumptions

Let us make a couple of assumptions.  First, we assume that we only need six different block types: G0_, H__, L__, R__, b__, M1x.  Here, x stands for a random number, meaning that the mover expands always at the same, but relative to other movers in the same machine at a random time.  Second, for encoding we will use the scheme depicted in Fig. 25 as discussed in [4].  It has the advantage that lock and key can reverse function.

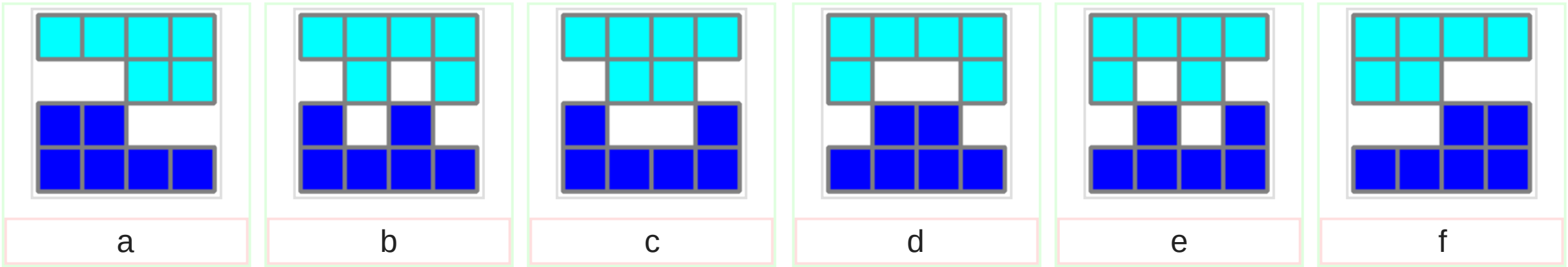

Figure 25: Key and lock

This type of encoding will only work for even codons, because it is essential that lock and key can reverse function.  In general there are C(n, n/2) combinations, where n is even.  For n=2 this results in two combinations, for n=4 in six, and for n=6 in twenty combinations.  If we restrict ourselves to the six different block types above, then n=4 suffices.

This translates into extrusions that should be slightly modified: they need to be slightly above and slightly below average, as indicated in Fig. 26.  For example, we can choose Fig. 26(a) for G0_, Fig. 26(b) for H__, Fig. 26(c) for L__, Fig. 26(d) for R__, Fig. 26(e) for M1x, and Fig. 26(f) for b__.

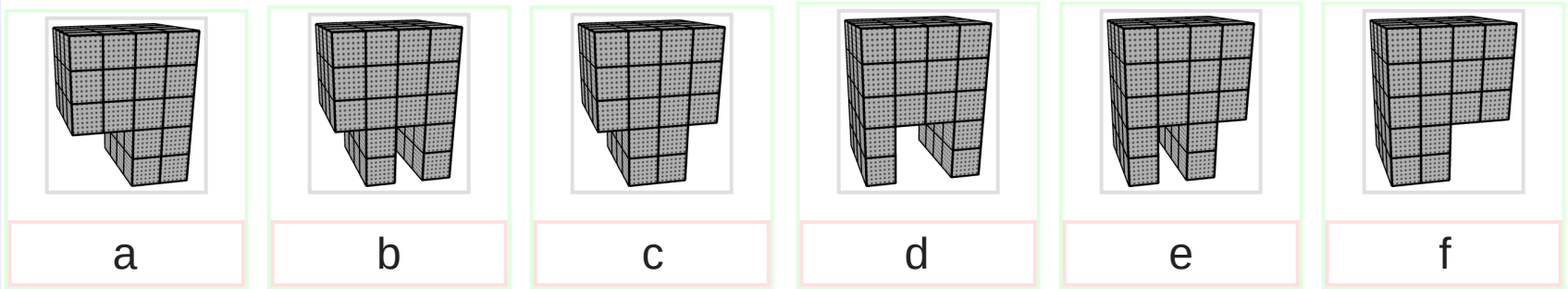

Figure 26: New markings

It is interesting to notice that the key-lock pairs (a,f), (b,e), and (c,d) are each others opposite.  Therefore, encodings are used for storing information, as well as for reading information.  Hence, if we were to start out with the yellow RNA tape, as indicated in Fig. 27, then matching the grey blocks would result in a negative copy of the tape.  Doing so twice results in a positive copy.  Now, the matching could occur purely randomly or via a machine.  In addition, this means that half of the machines are useless because they are negative copies folded.  They may or may not be doing anything that is useful.

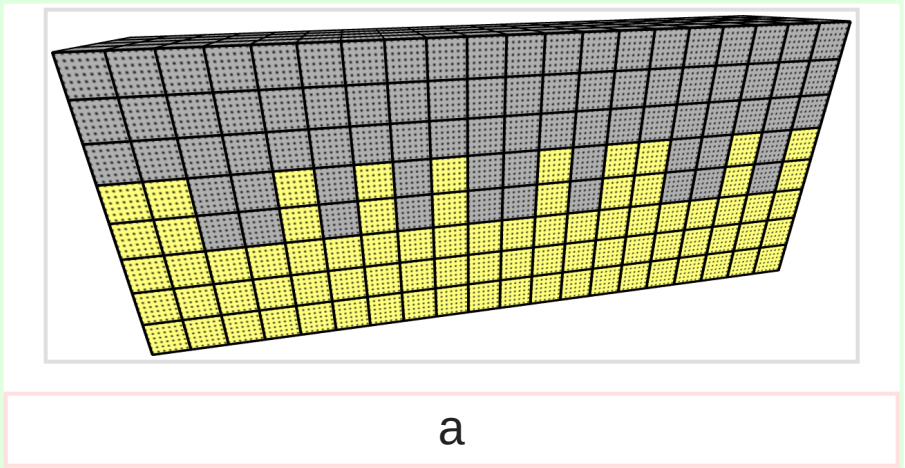

Figure 27: Matching/Copying

The chain cannot be folded during the matching/copying.  Thus, two things can happen here: if the folding takes a long time compared to copying, then many copies will be made before folding occurs.  Alternatively, if this is not the case, folding must be triggered by some mechanism, event, or catalyst.  Therefore, we first must make a copy and only then allow for the folding to occur.

## New Mechanism

As indicated, this copy process, that is the matching, could happen randomly by itself.  From an evolutionary point of view, this may actually have occured.  However, this is slow, and prone to errors, such as the blocks being slightly shifted to the left or right, etc.  To accelerate the process and reduce the likelihood of error, a dedicated machine is required.

This machine first selects a random block and attempts to match it to the current position. If it does not fit, the block is moved to the side. However, if it does fit, it leaves this block in place and moves the RNA tape by one, repeating the process.

Figure 28 shows a scetch of a machine that implements this algorithm. The RNA tape is depicted in white. A block randomly enters the volume below Mover1. Mover1 pushes down, trying to match this block with the tape. If it matches, it will fit nicely, otherwise, it will stick out a little. This is where Mover2 enters, it removes the block if it does not fit perfectly. Now, the third mover moves the whole RNA forward by one, but only if the random piece has fit in the previous step. This process repeats until the end of the RNA is reached. While pushing forward, the Gluer glues the pieces together. Thus, a negative copy of the RNA tape is created.

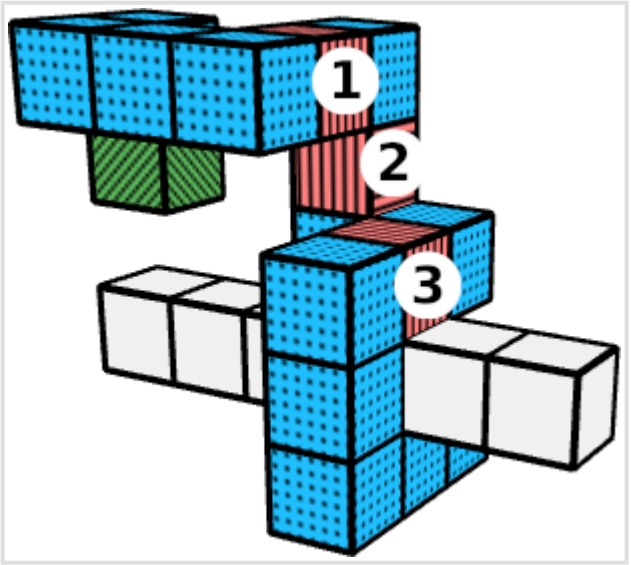

Figure 28: Universal Copier-Constructor

When attempting to implement this machine, there ensue several requirements regarding the details of the exact shape of the blocks. For demonstration purposes we assume a block to be made up of 4x4x4 subunits, but this is completely arbitrary: it could also be n x n x n, where n is larger than three. Because of the extrusion for the markings, all blocks will then have a volume of 4x4x5 subunits. This implies that the restriction that $x,y,z \in \mathbf{Z}$ is no longer true in general. However, overall, we would still like the positions of the blocks, at least within one machine, to be quantized.

Interestingly, this leads to severe restrictions on the shape of the Mover blocks. If we look at the RNA tape in Fig. 28, it must be such that the markings point upwards, otherwise they can not be read. This indicate that they are five subunits high. However, as we can see, the tape has to fit under the Mover3 block: at the same time, the tape has to be held in place when during the matching Mover1 pushes from above. This means that Mover blocks, at least partially, can only be three subunits high, so that the RNA fits beneath it. However, the volume of the Mover still needs to be of the general volume 4x4x5, and it cannot be 3x4x5 or 4x3x5. Hence, the 5th row must have a sparing for this on at least one side.

Furthermore, there are restrictions on the Mover blocks' internals, coming from the requirements on Mover2. After the matching, three things can occur: first, there is a match, the block to be matched will not stick out at all (Fig. 30(b)). Second, the block lies on its side, then the block to be matched will stick out by one subunit (Fig. 30(f)). Third, the block had exactly the opposite direction, then the block to be matched will stick out by two subunits (Fig. 30(g)). Only in the first case, Mover2 should not remove the block; in the other two cases, it should remove it. Figure 29 shows a Mover block that fulfills all of these requirements.

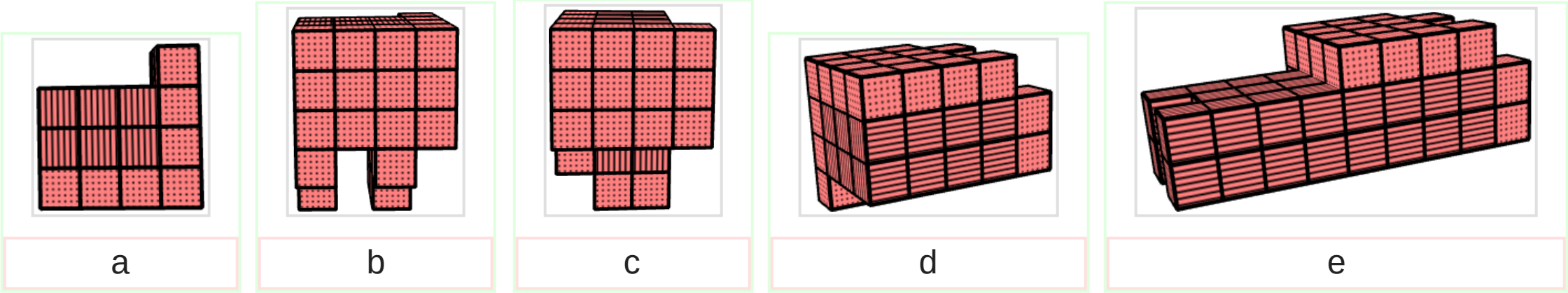

Figure 29: Detail new Mover block

A question to ask, why should the 5th row be shorter on both sides? When pushing down the block to be matched through Mover1, we have no way of telling which way they are oriented, upto a rotation of 180 degrees. If we had a presorter, we could, but we do not want a presorter.

Thus, if one of the blocks was rotated in the wrong way, the newly copied RNA would no longer smoothly fit in the copy machine, but would block it. However, if the 5th row is shorter on both sides, the newly copied RNA may have an encoding error (i.e. a mutation); however, it would still fit into the copy machine and not block it. Blocking is desastrous. The encoding error would result in the fact that we cannot distinguish (a,f) and (b,e), hence there will be mutations in making copies.

We are now ready to consider a detailed implementation of the universal copier-constructor (Fig. 30), focusing only on the functional blocks. In Fig. 30(a), the block to be matched (blue) has entered the volume below Matcher1. In this case, the block matches with the RNA, and Mover1 pushes it into the RNA tape (Fig. 30(b)). Mover2 then has no effect (Fig. 30(c)), and Mover3 moves the RNA tape forward by one (Fig. 30(d)). Figure 30(e) shows a block (blue) that has entered the volume below Matcher1 with the wrong orientation. When pushed down by Mover1, it sticks out by one subunit. In this case Mover2 will push the block to the side (Fig. 30(h)), and Mover3 has no effect on the RNA tape. Finally, Fig. 30(g) shows a block (blue) that has entered the volume below Matcher1 with the correct orientation, but it is of the wrong type. When pushed down by Mover1, it will stick out by two subunit. Again, Mover2 will push the block to the side (Fig. 30(h)), and Mover3 has no effect on the RNA tape.

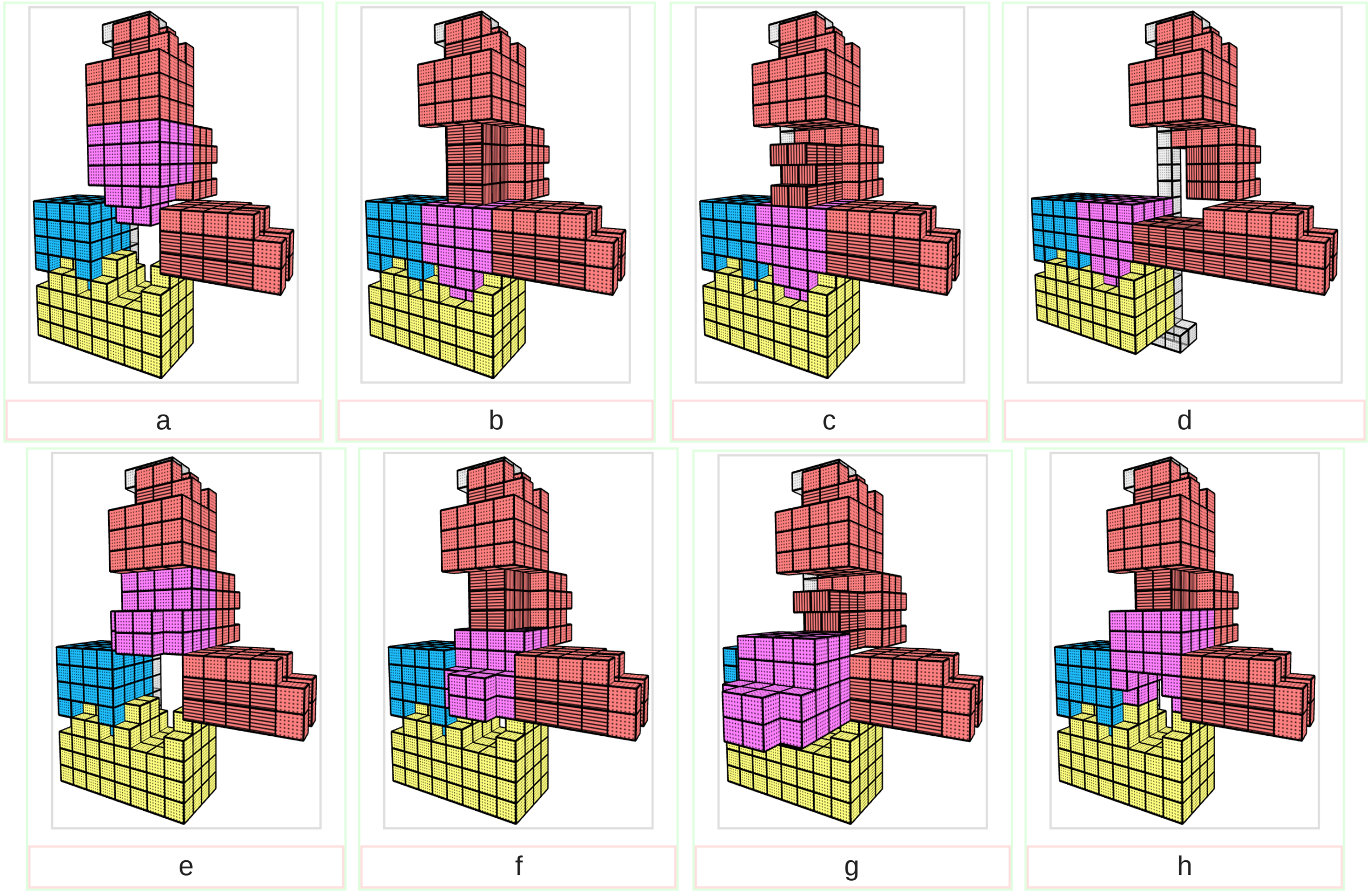

Figure 30: Detailed matching mechanism

.

Therefore, the interesting question is, can we create a machine like the one depicted in Fig. 30 via our folding mechanism, using the blocks G0_, H__, L__, R__, b__, M1x only. Figure 31 shows five different such machines created using the folding technique. Interestingly, each of them has a little quirk, which we leave the reader to discover.

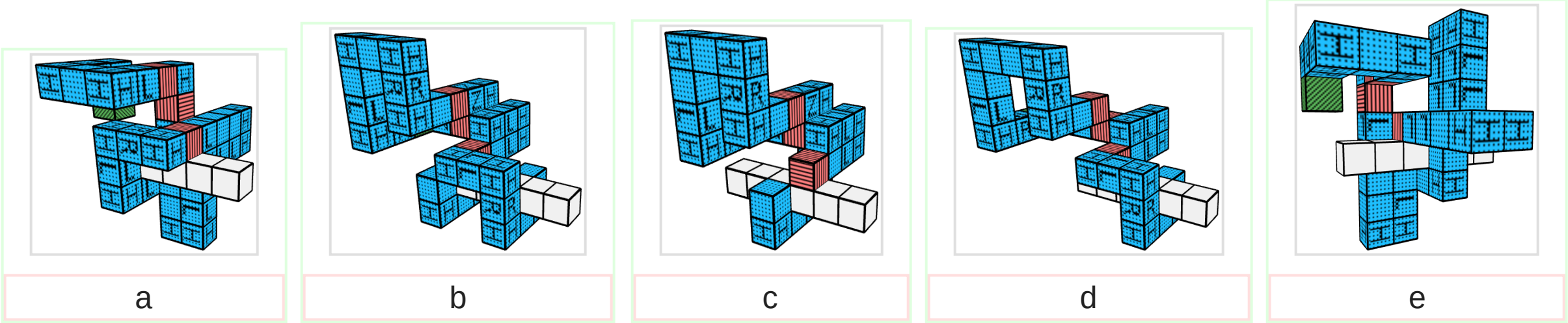

Figure 31: Universal Copier-Constructor created through folding

The lengths of these different machines vary between 33 and 40 blocks. Because we only have six block types, we can encode the information of one block with three bits; thus, our universal copier-constructor has approximately 100 to 120 bits of information content. In his famous paper "Information, reproduction, and the origin of life" [5] H. Jacobson made various assumptions and predictions about evolution, and one of these predictions was that the simplest living beings could have at most about 200 bits of information for evolution to ever occur in the time frame and conditions found on Earth.

Our universal Copier-Constructor is von Neumann's [6] automaton A and B in one machine. It has some interesting properties, such as it does not matter if the RNA is entering the wrong way, that is rotated around the length axis. The description of the machines (RNA or Phi in von Neumann's language) is now much shorter: it is same length as the machines themselves, O(n). However, at the fundamental level, this model cannot perform computations. It introduces two time scales: a time scale for copying and a time scale for folding. In addition, before our assumptions on the basic building blocks were absolutely minimal. Here what we do, we place many requirements on the basic building blocks, especially on the Mover. However, as von Neumann points out, we should not place to much logic/trickery into the basic building blocks. Each of these requirements on the shape and functioning of the blocks is equivalent to introducing a new little machine, as we saw with the presorter that would eliminate the equirement for the 5th row to be shorter on both sides.

.

## First Copier-Constructor

Of Sydney Brenner's [8] three questions, we have not answered the last question: How did it get that way? This is an evolutionary question. Initially, we would have no machines at all, but only randomly floating basic building blocks. Once in a while, through Brownian motion, some enter the area of influence of a Gluer block and might be glued together, thus creating larger random blobs. This is a highly inefficient process.

However, some of these random blobs might look similar to what we call the First Copier-Constructor, as depicted in Fig. 32. It consists of five blocks, "M1xH__b__b__G1_" or short MHbbG. Any block that enters the volume in front of the Mover will be pushed towards the Gluer block and glued to anything that might be there already.

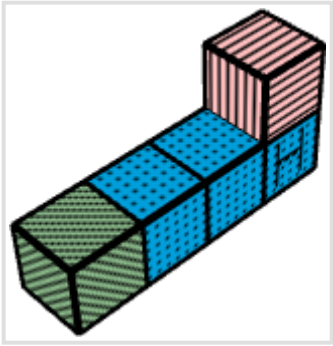

Figure 32: First Copier-Constructor: MHbbG

This is significantly more efficient for building random blobs than the pure random process. Moreover, if by chance it receives blocks in the order 'M', 'H', 'b', 'b', and 'G', it will actually create a copy of itself, with a likelihood of about one in a million.

Once the MHbbG has been created, we significantly increase the speed at which blobs are produced. Things are still created randomly but are much faster. This is reminiscent of Herbert A. Simon's Watchmaker Parable [7], where, as we climb up the evolutionary ladder, there are intermediary, stable steps. The simplicity of MHbbG, which requires only five blocks to create a copy of itself, highlights the feasibility of its random emergence. Despite the low probability, once the MHbbG is formed, the pace of evolution accelerates, setting the stage for further evolutionary steps and development.

.

## Discussion

The primary question that drove this research was why nature employs one-dimensional chains, such as peptides, to construct complex three-dimensional structures and machines such as proteins. Our findings suggest that the answer lies in the remarkable efficiency of this method.

This study demonstrates that it is surprisingly straightforward to replicate the protein-folding mechanism using our proposed approach. Although a few new block types are required, the overall number of necessary block types is significantly reduced, which is crucial because it directly affects the size of the codons. In addition, some machines, such as the Builder, become obsolete and further streamline the process. This results in a notable reduction in the amount of building material required and the RNA tapes of these machines become much shorter.

Two other key issues addressed in this study are the rotation and orientation challenges that plagued the previous mechanism. These issues have been resolved. Moreover, some of the functionalities that von Neumann delegated to automaton C, such as the transport problem, were effectively managed by the Track, Walker and Shuttle.

Evolutionary considerations are quite interesting. Brenner's last question [8], 'How did it get that way?' in the old model could not be answered. In the new model, with the steps of random self-copying, MHbbG, and then the universal copier-constructor, we have seen three steps on the evolutionary ladder that evolution can take before arriving at a more elaborate scheme.

This is also a question of reliability, how good are the copies that are being made, and how many errors/mutations do they contain? The simpler the mechanism, the more likely are errors, which are desirable from an evolutionary point of view.

In this study, we have completely neglected energy considerations. However, we do want to point out that the Mover blocks will undoubtedly require the most energy. As for Glue blocks, that depends on the exact mechanism. However, the folding and rotating blocks only require energy once. Similar to a loaded spring, this could have been provided to them during their creation. After unfolding, no additional energy is required.

.

## Conclusion

By introducing the concept of folding, this study illustrates that self-replication can be achieved with fewer machines, reduced coding requirements, and fewer building materials. Estimating the genome size for a minimal folding model, which includes components such as the Sorter, Recyclers, Decoder, and Copier, suggests that approximately 1,000 codons are needed. This is an order of magnitude less than that required for the previous 3D printing model introduced in [3], providing a clear answer to why nature favors one-dimensional peptide chains for building three-dimensional structures—because it is highly efficient.

In general, the previous method required encoding a volume, which grows as n^3. In contrast, the new method encodes a line, growing as n, or in the case of two-dimensional structures like β-sheets, it grows as n^2. This shift also allows the self-assembly of structures without the need for a Builder, as machines and structures can now self-assemble through folding.

The ability to rotate the chains introduces a significant improvement, as we can now control the final orientation of our machines, a capability which was previously lacking. In addition, the rotational degrees of freedom dramatically reduce the number of required block types. We only need one type of Gluer block instead of six, and the number of Mover blocks is reduced by a factor of three, whereas the number of Sorter blocks is halved.

In the old model, the Dissolvable block is necessary for positional information, serving as a placeholder. In the new model, this block addresses two other issues: it solves the problem of stop codons, providing a straightforward way to determine when a machine or structure is complete, and it enables the construction of quaternary structures, which are machines or structures composed of multiple pieces.

.

## Acknowledgements

I would like to sincerely thank L. Ochs for asking the right questions and insisting on the origami approach to the end. Without his persistence this work would not exist.

.

.

## Appendix

In the following, we describe all machines presented in this study in terms of their MDL.

Fig. 4a:

```
b__b__b__H__b__
```

Fig. 5a:

```
b__H__b__b__b__H__b__
```

Fig. 5b:

```
b__H__b__b__R__H__b__
```

Fig. 5c:

```
b__H__b__b__Z__H__b__
```

Fig. 5d:

```
b__H__b__b__L__H__b__
```

Fig. 7a:

```
b__b__H__L__H__L__H__L__H__L__H__L__H__L__H__L__H__L__H__L__
```

Fig. 7b:

```
b__b__H__R__H__R__H__R__H__R__H__R__H__R__H__R__H__R__H__R__
```

Fig. 7c:

```
b__b__H__Z__H__Z__H__Z__H__Z__H__Z__H__Z__H__Z__H__Z__H__Z__
```

Fig. 8a:

```
b__b__H__H__b__Z__b__H__H__b__Z__b__H__H__b__Z__b__H__H__b__
Z__b__
```

Fig. 8b:

```
b__b__H__b__H__b__Z__b__H__b__H__b__Z__b__H__b__H__b__Z__b__
H__b__H__b__Z__b__
```

Fig. 8c:

```
b__b__H__b__b__H__b__Z__b__H__b__b__H__b__Z__b__H__b__b__H__
b__Z__b__H__b__b__H__b__Z__b__
```

Fig. 9a:

```
b__h__H__H__h__
```

Fig. 9b:

```
b__H__h__b__b__
```

Fig. 9c:

```
b__h__H__H__h__b__
```

Fig. 9d:

```
b__h__H__H__h__h__H__
```

Fig. 9e:

```
b__H__h__h__H__b__H__h__h__H__H__h__
```

Fig. 9f:

```
b__H__h__h__H__b__H__h__b__b__
```

Fig. 9g:

```
b__H__h__h__H__b__H__h__h__H__b__
```

Fig. 9h:

```
b__H__h__h__H__b__H__h__h__H__
```

Fig. 11a:

```
G2_H__b__b__H__L__H__M24b__H__G2_G2_L__H__L__H__b__b__H__b__
R__H__M50M50M50M50R__
```

Fig. 11b:

```
G2_L__L__H__b__b__H__b__b__L__H__M24b__H__G2_G2_b__b__L__H__
L__b__H__b__b__H__L__L__L__H__L__L__M20M20M20M20L__
```

Fig. 13a:

M48b__H__H__R__H__b__b__M51b__H__b__b__b__b__b__b__b__L__H__
L__h__H__R__H__H__h__h__H__b__b__b__L__H__b__R__H__H__M32M32
M41

Fig. 13b:

G0_H__b__b__H__Z__M48b__H__H__R__H__b__b__M51b__H__b__b__b__
b__b__b__b__L__H__L__h__H__R__H__H__h__h__H__b__b__b__L__H__
b__R__H__H__M32M32M41

Fig. 15a:

b__b__H__R__b__b__b__b__b__H__b__b__b__H__H__L__H__H__Z__H__
M10R__H__b__b__H__L__b__H__Z__H__H__R__b__H__b__b__H__L__H__
R__H__b__b__b__b__b__

Fig. 15b:

b__b__H__R__b__H__H__Z__H__Z__b__b__b__b__H__b__b__b__H__H__
L__H__H__Z__H__M10R__H__b__b__H__L__b__H__Z__H__H__R__b__H__
b__b__H__L__H__R__H__b__b__b__b__b__

Fig. 15c:

b__b__H__R__b__b__b__b__b__H__b__b__b__H__H__L__H__H__Z__H__
M10R__H__b__b__H__L__b__H__Z__H__H__R__b__H__b__b__H__L__b__
H__H__Z__b__H__H__L__H__b__b__b__b__b__

Fig. 15d:

b__b__H__R__b__H__H__Z__H__Z__b__b__b__b__H__b__b__b__H__H__
L__H__H__Z__H__M10R__H__b__b__H__L__b__H__Z__H__H__R__b__H__
b__b__H__L__b__H__H__Z__b__H__H__L__H__b__b__b__b__b__

Fig. 16a:

M50M50M50d__h__d__d__M33M33H__H__R__H__b__h__b__h__b__S__R__
b__h__h__R__H__M47

Fig. 16b:

M50M50M50d__h__d__d__M33M33H__H__L__H__b__h__b__h__b__S__L__
b__h__h__L__H__M47

Fig. 17b:

M02M02M02M02H__H__b__b__b__b__b__b__b__b__L__G2_

Fig. 17c:

b__b__b__b__h__H__H__L__H__b__b__b__b__b__G1_b__h__L__h__h__

```
b__b__H__M51M51L__b__H__b__H__M34M34
```

Fig. 18a:

```
d__H__L__h__h__L__H__H__R__h__h__R__H__d__
```

Fig. 18b:

```
d__H__L__h__h__L__H__H__b__R__h__h__R__b__H__d__
```

Fig. 18c:

```
d__H__b__L__h__h__L__b__H__H__R__h__h__R__H__d__
```

Fig. 18d:

```
d__H__b__L__h__h__L__b__H__H__b__R__h__h__R__b__H__d__
```

Fig. 19:

```
G0_H__b__M10b__M10b__M10b__M10R__H__L__b__b__H__b__b__H__b__
b__R__H__b__H__L__b__b__H__b__H__L__H__b__H__R__H__b__H__b__
b__R__H__b__H__L__b__b__H__b__H__L__H__b__H__R__H__b__H__b__
b__b__b__
```

Fig. 20:

```
b__H__L__L__H__b__H__R__R__H__b__H__L__L__H__b__H__R__R__H__
L__H__R__b__H__H__L__H__b__d__M40b__M31M31M42
```

Fig. 21:

```
H__L__L__H__b__H__R__R__H__b__H__L__L__H__b__H__R__R__H__L__
H__R__b__H__H__L__H__b__d__M40b__M31M31M42H__M00M00R__H__R__
H__L__H__R__b__b__b__b__H__R__H__H__b__b__b__
```

Fig. 22a:

```
M07M07b__H__b__H__b__h__H__b__H__h__R__H__b__H__Z__H__b__H__
Z__H__b__H__Z__H__b__H__Z__H__b__H__Z__H__b__H__R__h__H__b__
H__h__b__H__b__H__b__M37M37
```

Fig. 22c:

```
M50M01M01b__M52b__b__H__H__L__h__H__b__b__b__b__b__H__h__L__
H__H__L__b__M42b__M31M31M40
```

Fig. 23:

```
b__L__L__H__b__b__b__H__L__H__R__H__M00M00H__L__d__
```

Fig. 31a:

```
G0_H__R__H__b__H__L__M10H__M13L__b__b__b__H__H__L__b__H__L__
L__H__H__b__b__L__M16H__R__H__L__H__L__
```

Fig. 31b:

```
b__H__b__b__H__b__b__b__H__Z__M40b__H__R__H__b__H__b__L__H__
R__G2_H__Z__M43H__L__H__H__L__b__M46H__L__H__R__H__b__H__b__
```

Fig. 31c:

```
b__H__b__b__H__b__b__b__H__Z__M40b__H__R__H__b__H__b__L__H__
R__G2_H__Z__M43H__L__H__H__L__b__M46
```

Fig. 31d:

```
M40b__H__R__H__b__b__H__b__L__H__R__G2_H__L__L__M43H__L__H__
H__L__b__M46H__L__H__R__H__b__H__b__
```

Fig. 31e:

```
G0_H__R__H__b__H__L__M10H__M13L__b__b__b__H__H__L__b__H__L__
L__H__H__b__b__L__M16L__L__b__H__H__R__H__L__L__H__H__L__L__
b__b__H__b__H__b__
```

Fig. 32:

```
M12H__b__b__G1_d__
```

.